\documentclass{article}

\usepackage{arxiv}

\usepackage[utf8]{inputenc} 
\usepackage[T1]{fontenc}    
\usepackage{hyperref}       
\usepackage{url}            
\usepackage{booktabs}       
\usepackage{amsfonts}       
\usepackage{nicefrac}       
\usepackage{microtype}      
\usepackage{lipsum}		
\usepackage{graphicx}
\usepackage{natbib}
\usepackage{doi}
\usepackage{amsmath}
\usepackage{caption}
\title{PIG-Nav: Key Insights for Pretrained Image Goal Navigation Models}

\date{} 					

\author{ 
    Jiansong Wan, Chengming Zhou, Jinkua Liu, Xiangge Huang, Xiaoyu Chen, Xiaohan Yi\textsuperscript{1}, \\[0.05cm]
    \textbf{Qisen Yang, Baiting Zhu, Xin-Qiang Cai, Lixing Liu, Rushuai Yang, Chuheng Zhang\textsuperscript{1}, } \\[0.05cm] \textbf{Sherif Abdelfattah, Hayong Shin, Pushi Zhang\textsuperscript{1}\thanks{Contact: \texttt{\href{mailto:pushizhang@microsoft.com}{pushizhang@microsoft.com}}. }\ , Li Zhao\textsuperscript{1}, Jiang Bian\textsuperscript{1}} \\[0.15cm]
    \textsuperscript{1}Microsoft Research
	 \\
}

\renewcommand{\undertitle}{}

\hypersetup{
pdftitle={A template for the arxiv style},
pdfsubject={q-bio.NC, q-bio.QM},
pdfauthor={David S.~Hippocampus, Elias D.~Striatum},
pdfkeywords={First keyword, Second keyword, More},
}

\begin{document}
\maketitle
 
\begin{abstract}

Recent studies have explored pretrained (foundation) models for vision-based robotic navigation, aiming to achieve generalizable navigation and positive transfer across diverse environments while enhancing zero-shot performance in unseen settings.

In this work, we introduce PIG-Nav~(Pretrained Image-Goal Navigation), a new approach that further investigates pretraining strategies for vision-based navigation models and contributes in two key areas. Model-wise, we identify two critical design choices that consistently improve the performance of pretrained navigation models: (1) integrating an early-fusion network structure to combine visual observations and goal images via appropriately pretrained Vision Transformer~(ViT) image encoder, and (2) introducing suitable auxiliary tasks to enhance global navigation representation learning, thus further improving navigation performance. Dataset-wise, we propose a novel data preprocessing pipeline for efficiently labeling large-scale game video datasets for navigation model training. We demonstrate that augmenting existing open navigation datasets with diverse gameplay videos improves model performance.

Our model achieves an average improvement of 22.6\% in zero-shot settings and a 37.5\% improvement in fine-tuning settings over existing visual navigation foundation models in two complex simulated environments and one real-world environment. These results advance the state-of-the-art in pretrained image-goal navigation models.  Notably, our model maintains competitive performance while requiring significantly less fine-tuning data, highlighting its potential for real-world deployment with minimal labeled supervision. 

\end{abstract}    
\begin{figure*}[t]
\centerline{\includegraphics[scale=0.5]{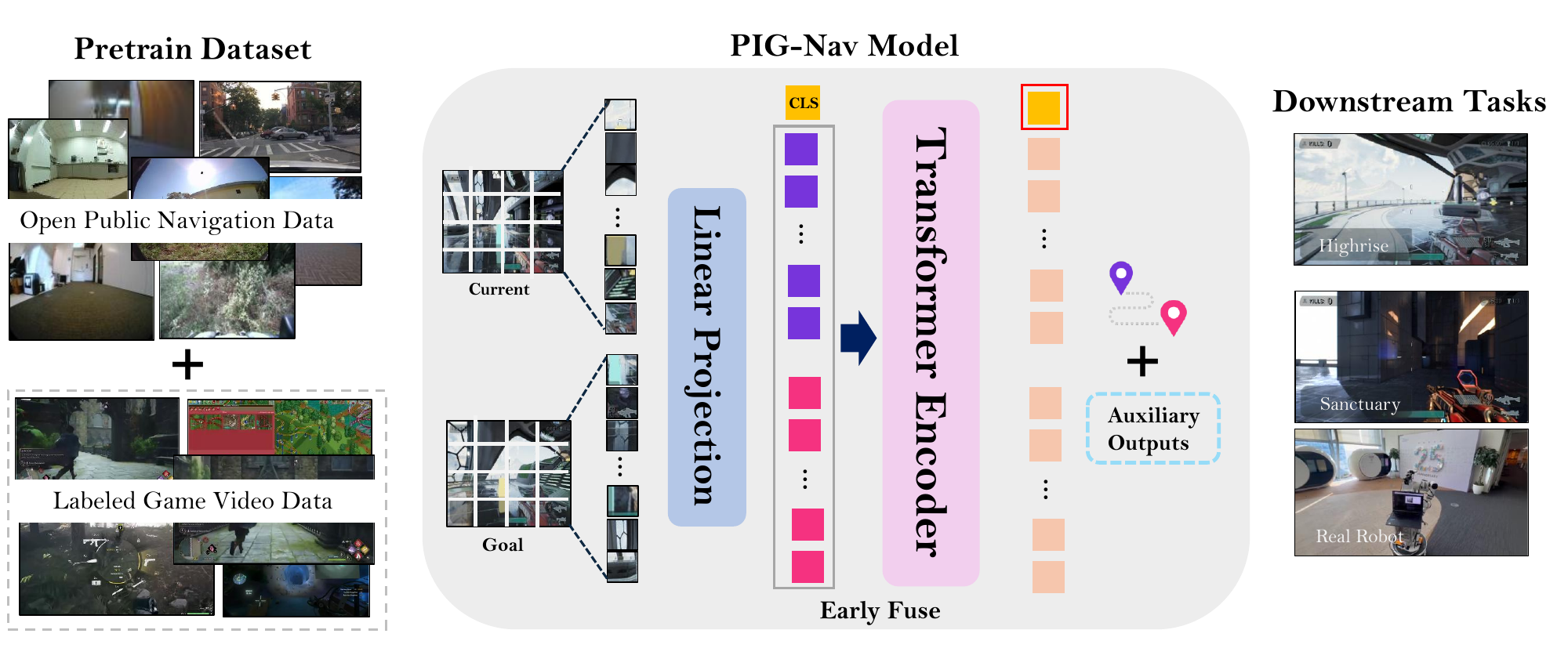}}
\caption{Overview of PIG-Nav.}
\label{fig1}
\end{figure*}

\vspace{-0.5mm}
\section{Introduction}
Vision-based navigation enables an autonomous agent to navigate toward a specific location or object using visual observations \citep{mayo2021visual, bonin2008visual}. This field has gained significant research attention due to its wide-ranging practical applications, including home automation, search-and-rescue missions, and last-mile delivery~\citep{shah2021ving, szot2021habitat, wu2022image, du2021vtnet}. 

Recently, researchers have explored pretrained visual navigation foundation models, which serve as generalized navigation frameworks capable of adapting to diverse robotic embodiments and environments~\citep{shah2023gnm, shah2023vint, sridhar2024nomad, zhang2024navid, zhang2024uni}. These models aim to capture prior navigational knowledge from large-scale navigation datasets \citep{karnan2022socially, liu2024botanicgarden, yokoyama2024hm3d}, allowing them to be deployed to new downstream tasks or fine-tuned with minimal data across different robotic platforms. Existing works have investigated unifying action spaces across various robot embodiments \citep{shah2023gnm, shah2023vint} and training generalized policies for both goal-directed navigation and environment exploration without explicit goal guidance \citep{sridhar2024nomad}.

However, despite these advancements, generalization to novel environments remains a major challenge, particularly when the new environments are out-of-distribution (OOD) compared to the pretraining dataset. Two key limitations contribute to this issue: 1) The design of existing models focuses on the relative information between observations and goals as a whole, failing to capture low-level correspondences. Additionally, they often employ low-capacity encoders that are not initialized with pretrained weights, restricting their ability to learn transferable navigation representations. 2) The datasets used for pretraining are not diverse enough to encompass a broad range of navigation scenarios, limiting the model's robustness.

To address these challenges, we introduce several key advancements to enhance the capabilities of visual navigation foundation models. First, we propose a simple yet effective model structure that enables fine-grained modeling of the low-level correspondence between images and goals. This is achieved by integrating a novel early-fusion network architecture and a pretrained Vision Transformer~(ViT) encoder to enhance representation learning. Second, we introduce auxiliary learning objectives that capture high-level navigation decision-making information, leading to improved representation learning and navigation performance. Furthermore, to address the substantial data requirements of training navigation foundation models, we propose a novel data labeling pipeline that efficiently processes game videos for navigation tasks, complementing existing navigation datasets.

To comprehensively evaluate PIG-Nav, we conduct experiments across two simulation-based game environments and a real-world robotic setting, comparing our approach against multiple baselines. Our results demonstrate that PIG-Nav exhibits superior generalizability, improved downstream fine-tuning performance, and strong potential as a foundation model for navigation tasks.

We demonstrate the following findings through our experiments: 
\begin{enumerate}
    \item PIG-Nav outperforms all baselines, including established visual navigation foundation models such as GNM \citep{shah2023gnm}, ViNT \citep{shah2023vint}, and NoMaD \citep{sridhar2024nomad}, 
    showing an average improvement of 22.6\%  in navigation success rate for zero-shot generalization settings, and an average improvement of 37.5\% for finetuning settings across all tasks.
    \item Pretraining PIG-Nav significantly enhances learning efficiency in downstream environments, reducing data requirements by approximately eightfold.
    \item The integration of an early-fusion network architecture with a pretrained Vision Transformer (ViT) leads to substantial improvements in navigation performance.
    \item Incorporating auxiliary tasks that capture global decision-making information further enhances the performance of the foundation model.
    \item The newly labeled game video dataset contributes to the training of navigation foundation models, demonstrating its potential as a valuable resource for future navigation tasks.
\end{enumerate}

\vspace{-0.5mm}
\section{Related Work}
\label{rw}

Image-goal navigation requires autonomous agents to navigate from their current location to a target location specified by a reference image.
This task is particularly challenging, as it requires agents to establish a robust correspondence between their current observations and the goal image.
When the current view does not directly reveal the goal, the agent must actively explore the environment to find relevant cues that guide it toward the goal location \citep{gupta2017cognitive, pertsch2020long, chaplot2020object}. 

Early methods primarily relied on handcrafted features to compare the goal image with the agent’s current view and estimate the direction toward the target.
However, these approaches often struggled in complex or dynamic environments, as they were highly sensitive to lighting variations, occlusions, and scene changes. To overcome these limitations, \citep{kwon2023renderable} proposed constructing a 3D map of the environment and projecting the goal image onto this map to compute the optimal navigation path \citep{VisualGraphMemory_kwon_2021}.

To overcome these limitations, recent research has embraced Deep Reinforcement Learning (DRL)~\citep{yadav2023ovrl, sun2024fgprompt, kim2023topological, oquab2023dinov2}.
DRL-based methods learn end-to-end navigation policies by interacting with simulated environments, using visual observations and goal images to make real-time decisions \citep{zhu2017target, shi2019end}.
However, due to the heavy reliance on simulation environments and overfitting issues associated with DRL-based approaches, there is a growing interest in exploring general pretrained navigation models \citep{StudyOverfittingDeep_zhang_2018, LearningTransferableVisual_radford_2021}.
These models leverage large-scale, diverse datasets and high-capacity architectures to achieve robust performance in unseen environments. GNM create a unified model that can operate across different robots and settings, achieving zero-shot deployment.
In contrast, ViNT leverages Transformer architecture to fuse features from the current view and the goal image at an early stage, which enhances the learning of fine-grained visual correspondences.
Building on ViNT, NoMaD introduces a masked diffusion strategy within a compact Transformer framework, leading to improved performance with lower computational overhead \citep{LearningTransferableVisual_radford_2021}.
These enhancements aim to improve generalization, data efficiency, and navigation performance in unseen environments.

Our work builds on these advancements by proposing a Vision Transformer (ViT)-based early fusion network that tightly integrates visual features from both the current observation and the goal image.
We further augment our architecture with auxiliary tasks to capture global decision-making information, and introduce a novel game video labeling pipeline to address the data requirements. This combination aims to enhance both generalization and data efficiency.

Another active research direction leverages Vision-Language Models (VLM) \citep{TransformersImageGoalNavigation_pelluri_2024}, which utilize VLM as a backbone to process both visual inputs and language instructions, guiding agent navigation through multi-modal reasoning ~\citep{NavGPT2UnleashingNavigational_zhou_2024,chen2024affordances, zhang2024uni, anderson2018vision, ContrastiveLocalizedLanguageImage_chen_2025, eftekhar2024one}.
These approaches differ from our focus.
However, our findings offer valuable insights that could later benefit multi-modal networks.

\vspace{-0.5mm}
\section{Proposed Model: PIG-Nav}
\label{method}
In this section, we provide a comprehensive explanation of our proposed model, PIG-Nav, including its architecture and training methodology, which incorporates auxiliary training tasks. Figure~\ref{fig1} presents an overview of our approach.

Our work begins with the collection of diverse trajectory data, consisting of RGB images along with their corresponding position and orientation information. Additionally, we annotate large-scale game videos to further enhance PIG-Nav’s capability. Using these datasets, we pretrain PIG-Nav to learn generalizable navigation representations, allowing it to adapt effectively to a wide range of navigation tasks. After pretraining, we fine-tune the model on environment-specific datasets, optimizing its performance for downstream applications.

Specifically, Section~\ref{section-model-1} introduces the overall architecture of PIG-Nav, while Section~\ref{section-model-2} details the training methodology, including the primary training task and the design of auxiliary tasks. We will elaborate on pertaining datasets in Section~\ref{section-dataset}. 

\vspace{-0.5mm}
\subsection{Model Architecture}
\label{section-model-1}

Figure~\ref{fig1} illustrates the network architecture of our model. The visual encoder is designed to capture correspondences between the low-level features of the observation and goal images, facilitating effective feature alignment. The current observation and the goal image are first independently processed into patch embeddings before being fused into a unified set of visual tokens. Each patch token is augmented with learnable tokens indicating the observation or goal images, which are then jointly fed into the Vision Transformer (ViT) encoder. We refer to this design as an early-fusion model architecture, as the transformer encoder directly operates on the low-level features of both the observation and goal images, enabling more effective spatial and semantic integration.

The learnable [CLS] token serves as an additional input to the transformer encoder,  enabling the extraction of joint representations from the image inputs. The resulting output embedding of this token is subsequently fed into separate MLP layers as the contextual embeddings to generate waypoint action outputs and auxiliary outputs. The action heads decode the next $N_{\text{waypoint}}=10$ actions, utilizing a 4-dimensional waypoint action space $[\Delta x, \Delta y, \cos \Delta\psi, \sin\Delta\psi]$, where $\Delta x, \Delta y, \Delta\psi$ are the 2D positional movements and rotations of the agent. The auxiliary prediction tasks are detailed in Section~\ref{section-model-2}. 

We also propose to leverage ViT encoders that pretrained by powerful self-supervised models including DINOv2~\citep{oquab2023dinov2} and Masked Auto Encoding~(MAE)~\citep{he2022masked}, enabling effective joint processing of observation and goal images. Further discussions on the effects of these pretrained ViT encoders can be found in Section~\ref{section-ablation}.

\vspace{-0.5mm}
\subsection{Training Tasks}
\label{section-model-2}

The primary objective of our model is to predict waypoint actions, enabling agents to determine how to move from their current locations. Additionally, PIG-Nav incorporates auxiliary tasks that provide high-level supervisory signals, guiding low-level navigation actions and leading to more effective trajectory planning. Specifically, we introduce three types of auxiliary tasks: 1) relative pose to the goal position, 2) distance to the goal position, and 3) global path to the goal position. Further calculations and details are provided in the Appendix.

\vspace{-0.5mm}
\paragraph{Waypoint Action Prediction.} The main task of PIG-Nav is to predict translation and rotation changes over the next $N_{\text{waypoint}} = 10$ consecutive steps in the provided navigation trajectory based on the current image observation and the goal image observation. This chunk-based action prediction approach has been adopted in recent works~\citep{bachmann2024pitfalls, hu2024learning} as an improvement over methods that predict only the next immediate action. 

\vspace{-0.5mm}
\paragraph{Relative Pose to Goal. }
In the relative pose prediction task, the model estimates the pose of the goal image relative to the observation image. This task enables the model to capture global task information by understanding the spatial relationship between the current observation and the goal image. 

\vspace{-0.5mm}
\paragraph{Navigation Distance Prediction. }
In the navigation distance prediction task, our model is trained to estimate the total distance required for the agent to navigate from its current state to the goal state. By training to predict the distance, the model learns to predict the connectivity and traversability between different locations in the environment, enabling more efficient path planning. 

\vspace{-0.5mm}
\paragraph{Global Path Prediction. }
To further enhance the agent’s ability to predict long-range paths, we introduce the global path prediction task. This task requires the model to predict $N_{\text{global}} = 10$ intermediate waypoints, which are equally spaced in time from the current timestep to the goal image's timestep. 
\section{Pretraining Datasets for PIG-Nav}
\label{section-dataset}

\subsection{Open Public Navigation Datasets}
\label{section-navigation-dataset}
We collect a large-scale dataset of real-world navigation trajectories to pretrain our model including GoStanford~\citep{hirose2019deep}, RECON~\citep{shah2021rapid}, CoryHall\citep{kahn2018self}, Berkeley DeepDrive~\citep{yu2020bdd100k}, SCAND~\citep{karnan2022socially}, TartanDrive~\citep{triest2022tartandrive}, and SACSoN~\citep{hirose2023sacson}. This combined dataset encompasses both teleoperated and autonomous navigation behaviors, collected from various robotic platforms. Additionally, it captures a wide range of motion dynamics, as the data originates from different robots operating in diverse environments, including office buildings, off-road trails, university campuses, and indoor rooms.

To ensure data quality and facilitate stable training, we filtered out data points where position changes exceeded five times the average displacement. Additionally, we follow previous works to transform the data into a unified space, as datasets vary in scale and properties. Further details regarding open public navigation datasets are provided in the Appendix.

\subsection{Labeled Navigation Datasets from Game Videos}

To further enhance the generalization and robustness of our PIG-Nav model from a data-centric perspective, we collect a large and diverse set of 3D game video data. We develop a data processing pipeline that includes preprocessing, filtering, and labeling to ensure high-quality training data for model pretraining.

\begin{figure*}[htb]
\centerline{\includegraphics[scale=0.4]{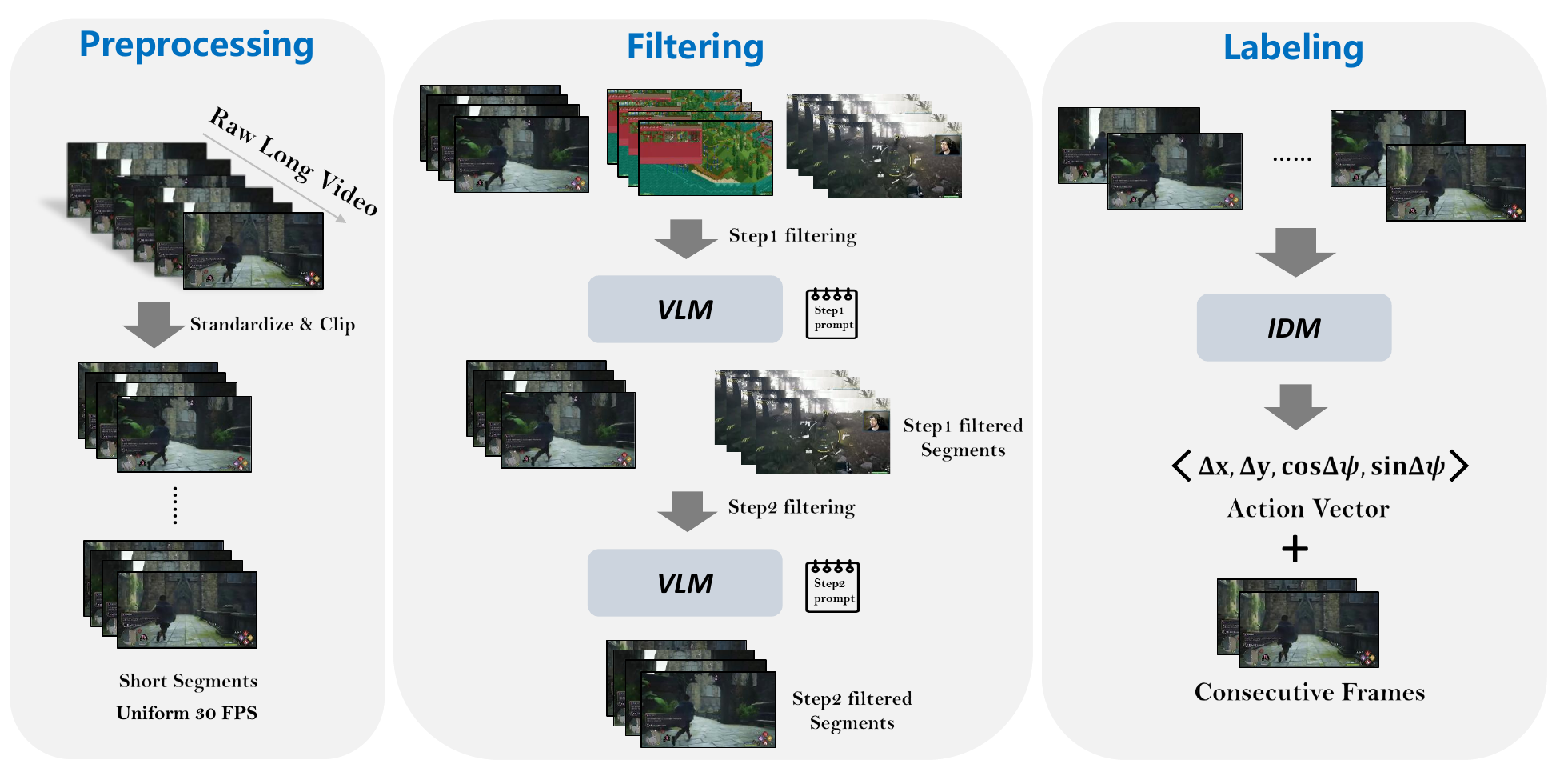}}
\caption{Game Video Data Processing Pipeline.}
\label{fig4}
\end{figure*}
\label{method}

Figure~\ref{fig4} illustrates our data cleaning and annotation pipeline, which consists of three key stages: preprocessing, cleaning, and labeling. The details of each stage are described below.

\vspace{-0.5mm}
\paragraph{Data Source \& Pre-processing. } 
We collect over 220,000 long gameplay videos, sourced from approximately 2,000 different games, as our original dataset. The average duration of these videos is around 3 hours. To facilitate processing, we segment the original videos into 10-second clips and standardize them to 30 FPS for subsequent processing.

\vspace{-0.5mm}
\paragraph{Data Cleaning by VLM.}

We utilize the open-source vision-language model Qwen2-VL-7B~\citep{wang2024qwen2} to classify whether the preprocessed video segments are suitable for inclusion in the navigation dataset. The Qwen2 model takes both video segments and language prompts as input and outputs a binary classification result: \textit{True} (suitable) or \textit{False} (unsuitable), based on the provided prompt instructions.

To ensure high-quality video selection, we apply a two-step filtering process and evaluate the precision, recall, and F1-score at each step. For the first filtering step, we sample total 500 videos from raw video segments randomly and manually label them as \textit{True} or \textit{False}. The Qwen2 model's initial classification achieves 52\% precision and 75\% recall, indicating that the first filtering step effectively retains most useful segments, but the precision remains relatively low, suggesting the necessity for further refinement.  

In the second filtering process, we experiment with 12 different prompt instructions and select the one yielding the best weighted balance between precision and recall to improve dataset quality. The final prompts we picked result in 85\% precision and 68\% recall, demonstrating that our two-step filtering strategy effectively constructs a high-quality video dataset for navigation tasks. All prompts we use in the data cleaning process and the statistics are detailed in Appendix.

\vspace{-0.5mm}
\paragraph{Action Annotation.} 

To annotate navigation actions $[\Delta x, \Delta y, \cos \Delta\psi, \sin\Delta\psi]$ from our filtered video segments, we train an Inverse Dynamics Model~(IDM) to generate action labels, which measures the action change between two consecutive video frames. The IDM adopts the same network architecture as PIG-Nav, excluding the waypoint acion and auxiliary outputs, and operates by processing two consecutive video frames that capture slight changes in motion. It predicts the corresponding navigation action between these frames using a dedicated action prediction head.

We train the IDM using open navigation datasets described in Section~\ref{section-navigation-dataset}. To optimize the accuracy of navigation action labels, we experiment with different input frame rates on small dataset batches. Based on these evaluations, we select a configuration where 48 short frame pairs are extracted from each 10-second filtered video for training, and the IDM is trained for 26 hours. Further details regarding the visualizations of labeled video segments can be found in the Appendix.
\section{Experiments}

We design our experiments to address the following key research questions:
\begin{itemize}
\item How does PIG-Nav compare to other pretrained navigation foundation models in terms of zero-shot generalization and fine-tuning performance in simulation and real world environments?
\item How does pretraining phase of PIG-Nav enhance the data efficiency of fine-tuning in unseen new environments?
\item Does the game video dataset, labeled using our proposed pipeline, improve the pretraining of navigation foundation models? 
\end{itemize}
Additionally, to further justify our design choices and provide insight into developing navigation foundation models, we conduct extensive ablation studies focusing on the following aspects:
\begin{itemize}
    \item Effect of early fusion – How does early fusion of low-level image features from the observation and goal images impact PIG-Nav's performance?
    \item Impact of pretrained ViT initialization – How do self-supervised ViT encoders (e.g., MAE, DINOv2) and ViT model size affect PIG-Nav's performance?
    \item Influence of auxiliary tasks – How do auxiliary tasks for predicting global decision-making information contribute to PIG-Nav’s effectiveness?
\end{itemize}
Next, we present the details of our environments used for testing our models and the experiment setup, and then present the result analysis. 

\subsection{Experiment Setups}

\begin{table*}[t]
\centering
\resizebox{0.9\linewidth}{!}{
\begin{tabular}{l|cccccc|cccccc|cc}
\toprule
               & \multicolumn{6}{c|}{Highrise}                                      & \multicolumn{6}{c|}{Sanctuary}                          & \multicolumn{2}{c}{Robot}  \\                  
               & \multicolumn{2}{c}{Easy} & \multicolumn{2}{c}{Medium} & \multicolumn{2}{c|}{Hard} & \multicolumn{2}{c}{Easy} & \multicolumn{2}{c}{Medium} & \multicolumn{2}{c|}{Hard} 
               & \multicolumn{2}{c}{8 Tasks}\\ 
               & SR         & SPL         & SR         & SPL         & SR         & SPL        & SR         & SPL         & SR         & SPL         & SR         & SPL  
               & SR         & SPL
               \\ \hline
GNM (ZS)  &  0.86          &     0.78        &     0.58       &     0.47        &      0.32      &      0.20      &     0.33       &  0.27           &      0.11       &      0.05       &               0.00  & 0.00  &  0.00  &  0.00   \\
    ViNT (ZS)  &   0.82     &   0.78      &   0.46     &    0.43     &   0.26     &   0.22     &      0.41      &      0.37       &      0.36      &      0.32       &     0.18 &      0.17      &  0.13  &  0.10 \\
    NoMad (ZS) & 0.78 & 0.75 & 0.42 & 0.40 & 0.22 & 0.19 & 0.40 &  0.31 & 0.42 & 0.36 &  0.08 & 0.05 & 0.00 & 0.00 \\
PIG-Nav (ZS)  &      \textbf{0.90}        &     \textbf{0.86}        &  \textbf{0.72}          &    \textbf{0.68}         &          \textbf{0.46}  &     \textbf{0.38}       &   \textbf{0.84}         &      \textbf{0.80}       &      \textbf{0.48}      &      \textbf{0.45}      &   \textbf{0.30} & \textbf{0.26}  & \textbf{0.50}  &    \textbf{0.44}   \\
\hline
GNM (FT)  &    0.98&     0.92&   0.68&   0.63&     0.66       &    0.55        &       0.72     &  0.67           &      0.38      &   0.32          &              0.20 & 0.15   & 0.25  &  0.22  \\ 
ViNT  (FT)  &    0.88    &    0.83    &   0.66     &   0.56     &    0.50    &    0.42   &       0.62      &      0.55        &       0.38     &       0.35       &     0.19 & 0.16   &  0.25  &   0.21    \\ 
NoMaD  (FT)  &   0.70    &   0.68     &    0.34    &  0.33     &     0.15   &   0.14    &     0.52       &       0.47       &      0.18      &      0.17        &   0.10   &  0.09    &    0.38  &    0.32  \\ 
PIG-Nav  (FT) &    \textbf{1.00}        &       \textbf{0.97}      &  \textbf{1.00}          &      \textbf{0.94}       &        \textbf{0.90}    &     \textbf{0.81}       &   \textbf{1.00}         &   \textbf{0.96}          &    \textbf{0.84}        & \textbf{0.75}            &    \textbf{0.76} & \textbf{0.68} & \textbf{1.00} & \textbf{0.89}  \\         
\bottomrule
\end{tabular}
}
\caption{SR and SPL performance across different environments and settings. All values are rounded to the nearest hundredth.}
\label{tab:performance_comparison}
\end{table*}

\vspace{-0.5mm}
\paragraph{Environment.}
We conducted evaluation experiments across two simulation environments and one real-world robotic environment. The two simulation environments, Highrise and Sanctuary, are both part of ShooterGame, a representative PC multiplayer First-Person Shooter (FPS) game built on Unreal Engine 4. This game provides a robust simulation framework for evaluating navigation models. Both environments span approximately 10,000 m², featuring large-scale, dynamic environments with numerous obstacles and complex background elements, making them particularly challenging for navigation policies.

For the real-world environment, we deploy our model on a wheeled robot traversing an indoor floor space of approximately 2,000 m², providing a realistic setting to evaluate the model’s performance in physical environments.

\vspace{-0.5mm}
\paragraph{Fine-tune Dataset. } We collect 2,100 navigation episodes by human players from Highrise and Sanctuary as our downstream task dataset. The average episode length is 34 steps in Highrise and 47 steps in Sanctuary, with approximately 0.7 million and 0.9 million total frames, respectively. For the real-robot dataset, we gather 113 episodes using human teleoperators, with an average episode length of 93 steps and a total of 10,500 frames. In all datasets, human players and teleoperators are instructed to traverse the entire environment, and we subsequently trimmed the trajectories to extract navigation episodes following optimal paths. For all three environments—Highrise, Sanctuary, and the real-robot setting—we split the dataset into 90\%  for fine-tuning dataset and 10\% for validation.

\vspace{-0.5mm}
\paragraph{Final Evaluation Task.}
 We randomly sample the start and goal points for evaluation within the corresponding environment. To assess performance under varying conditions, we divide the evaluation into three difficulty levels—Easy, Medium, and Hard— determined by the trajectory length between start point and goal point on ShooterGame. In Highrise, the average values of timesteps required to reach the goal for Easy, Medium, and Hard difficulty levels are  $(19, 37, 55)$ respectively. Similarly, in Sanctuary, the corresponding values are  $(28, 56, 83)$ timesteps. We sample 50 tasks for each difficulty level in Highrise and Sanctuary. For the robot experiments, we test a total of 8 tasks where the start and goal positions are distributed across the entire indoor floor. The agents are required to reach the goal from their current position within a total of 100 steps.

 \vspace{-0.5mm}
\paragraph{Evaluation Metrics.}
We employ two widely used navigation metrics: Success Rate (SR) and Success Weighted by Path Length (SPL), which assess both task completion and navigation efficiency.
We define these metrics as:
\begin{equation*}
SR = \frac{1}{N} \sum_{i=1}^N S_i, \quad SPL = \frac{1}{N}\sum_{i=1}^N S_i\frac{d_i}{p_i}
\end{equation*}
where $S_i$ equals 1 if navigation is successful in $i$-th evaluation episode and 0 otherwise, $N$ is the total number of evaluation episodes, $d_i$ represents the shortest distance from the starting position to the goal., and $p_i$ denotes the actual path length taken by the agent during evaluation. 

\vspace{-0.5mm}
\paragraph{Hyperparameter Settings.}
Our model utilizes ViT-Base as the image encoder and incorporates multiple MLP heads, resulting in a total of approximately 100 million parameters.

We train the model using the Adam optimizer with a learning rate of $5\times 10^{-5}$ and a batch size of 128. The pretraining phase spans 200 epochs, requiring approximately 2 days on a 4× A100 GPU setup. For fine-tuning, the model is trained for 200 epochs, completing in approximately two days on a single A100 GPU.

\begin{figure*}[htb]
\centerline{\includegraphics[scale=0.40]{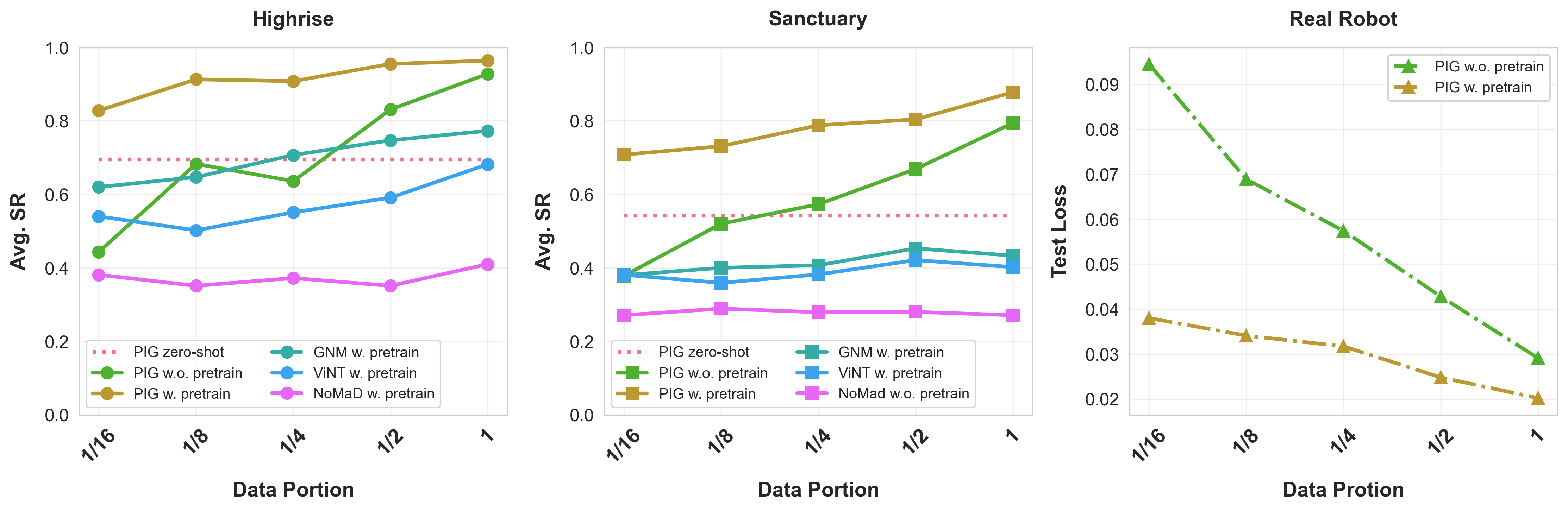}}
\caption{Average SR and Validation Loss performance comparison across different fine-tuning dataset size. Left for Highrise, middle for Sanctuary and right for real robot. Average SR is used for Highrise and Sanctuary measurement, with higher value meaning better model performance. Test Loss is used for real robot measurement, with lower value meaning better model performance.}
\label{fig:label5}
\end{figure*}
\label{method}

\subsection{Results \& Analysis}

\subsubsection{Comparison with Baselines for Zero-Shot Generalization and Finetuning}
In this experiment, we compare PIG-Nav against three baseline methods—GNM, ViNT, and NoMaD—across two key evaluation criteria: generalizability to unseen environments and fine-tuning performance in diverse settings. The results, presented in Table~\ref{tab:performance_comparison}, demonstrate that PIG-Nav consistently outperforms all three baselines in both zero-shot and fine-tuned tasks.

Notably, at the Medium and Hard difficulty levels, PIG-Nav surpasses the baselines by an even larger margin. We attribute this to PIG-Nav's effective model architecture design. The three baselines rely on CNNs to process the concatenated observation and goal images, capturing the relative information as a whole. In contrast, PIG-Nav directly models the spatial relationships among image patches via our designed early-fusion ViT, enabling it to more effectively capture the nuanced connections between the current observation and the goal image. This enhanced contextual understanding allows PIG-Nav to make more informed navigation decisions, particularly in challenging scenarios.

\begin{figure*}[htb]
\centerline{\includegraphics[scale=0.5]{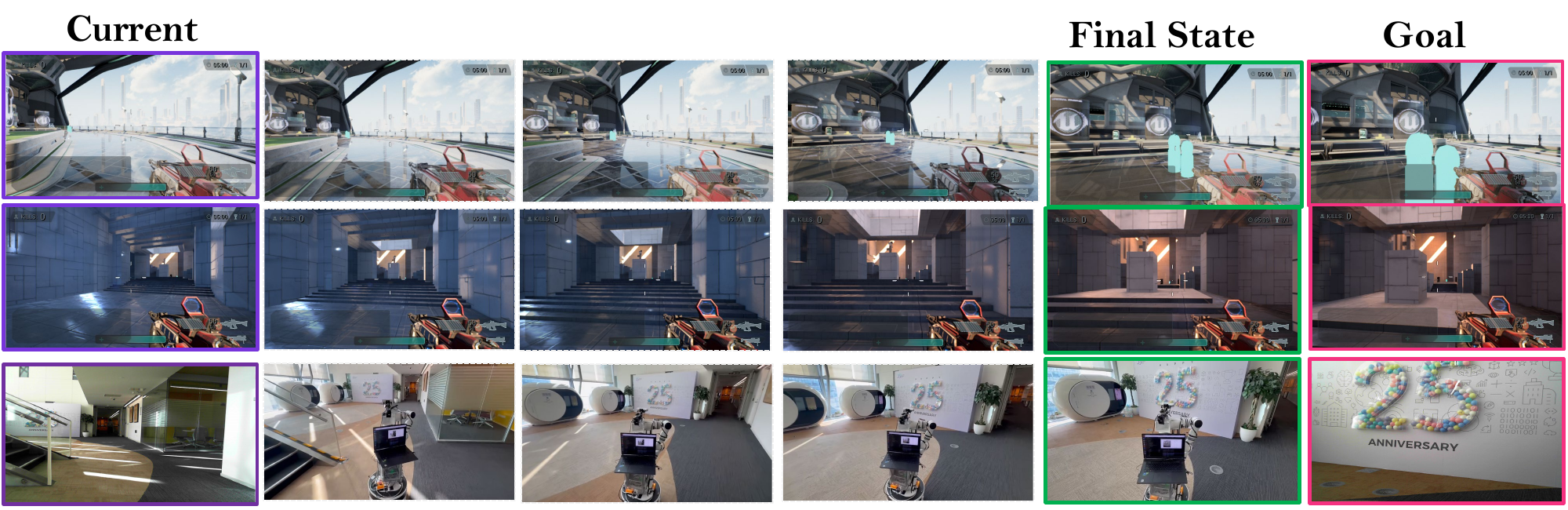}}
\caption{Three evaluation examples of PIG-Nav on three environments: Highrise, Sanctuary, Real Robot.}
\label{fig:label6}
\end{figure*}

We present three examples from each environment's evaluation test in Figure~\ref{fig:label6}. For more rollout examples, please refer to the Appendix.

\subsubsection{Fine-Tune PIG-Nav on New Environment}

In this experiment, we evaluate the fine-tuning efficiency across the three environments. To assess adaptability with limited labeled data, we progressively reduce the amount of fine-tuning data to different levels: $1$, $1/2$, $1/4$, $1/8$, and $1/16$ of the full dataset. We then measure the model's performance under each setting, averaging results across three difficulty levels.

As illustrated in Figure~\ref{fig:label5}, performance decreases across nearly all models as the available fine-tuning data is reduced. However, PIG-Nav, with its pretraining phase, demonstrates the most robust performance, maintaining high accuracy even with significantly less fine-tuning data compared to other baselines and PIG-Nav without pretraining. Notably, it reduces data requirements by approximately eightfold, further reinforcing its effectiveness as a navigation foundation model.

Additionally, it is worth highlighting that the zero-shot generalization performance of PIG-Nav surpasses models fine-tuned with $1/16$ and $1/8$ of the data without pretraining, underscoring its strong generalization capabilities, even in the absence of additional fine-tuning data.

\vspace{-0.5mm}
\subsubsection{Benefit of Labeled Game Video Data}
To assess the impact of our newly labeled game video data, we pretrain PIG-Nav using both this dataset and standard public navigation datasets and compare the average validation loss. As illustrated in Table~\ref{tab:gamevideo}, we find that incorporating additional labeled game video data results in lower validation losses, demonstrating the efficiency of our labeled dataset and its processing pipeline.

\begin{table}[h]
\centering

\resizebox{0.43\linewidth}{!}{
\begin{tabular}{c|c}
\toprule
& Average Validation Loss \\
\hline 
w/o Game Data & 0.189 \\ 
\hline
w/ Game data & \textbf{0.164} \\
\bottomrule
\end{tabular}
}
\caption{Loss comparison with and without Game Video Data.}
\label{tab:gamevideo}
\end{table}

\vspace{-0.5mm}
\subsubsection{Ablation Studies}
\label{section-ablation}

\paragraph{Early Fuse VS. Non-Early Fuse.}
To verify the effectiveness of our proposed early-fusion structure, we experiment with two versions of PIG-Nav: early-fusion and non-early-fusion, both combined with an MAE-pretrained ViT image encoder, in the Highrise and Sanctuary environments. The key difference between early-fusion and non-early-fusion is that early-fusion allows cross-attention between the observation and goal images, whereas non-early-fusion processes them separately and later fuses them by contrasting their learned [CLS] tokens from the ViT encoder.

Table~\ref{tab:earlyfuse} presents the performance comparison between these two structures. The results indicate that the early-fusion network architecture achieves better navigation performance in both zero-shot and fine-tuning settings, suggesting that early fusion of low-level features from observation and goal images enhances performance in image-goal navigation.

\begin{table}[h]
\centering
\resizebox{0.5\linewidth}{!}{
\begin{tabular}{l|cc|cc}
\toprule
               & \multicolumn{2}{c|}{Highrise}                                      & \multicolumn{2}{c}{Sanctuary}    \\                    
               & SR         & SPL         & SR         & SPL    \\ \hline
Non-Fuse (ZS) &     0.487 & 0.443      &     0.280 & 0.217  \\
Early Fuse (ZS) &  \textbf{0.693}    &  \textbf{0.640}     &  \textbf{0.540}    &  \textbf{0.503} \\
\hline 
Non-Fuse (FT) &  0.927    &  0.860     &  0.853  &  0.770 \\
Early Fuse (FT) &  \textbf{0.967}   & \textbf{0.907}     &   \textbf{0.867}   &  \textbf{0.797} \\
\bottomrule
\end{tabular}
}
\caption{SR and SPL performance on two types of model structure  across two environments.  Experiments run for zero-shot (first two rows) and fine-tuning (last two rows) settings. 
Performances are averaged on three difficulty levels. 
}
\label{tab:earlyfuse}
\end{table}

\paragraph{Integration of Pretrained Visual Encoders.}
We also tested the effect of various pretrained ViT encoders on PIG-Nav's performance, with the results presented in Table~\ref{tab:vit}.

The experimental results show that ViT pretrained with Self-Supervised Learning (SSL) methods significantly outperforms models trained from scratch. Additionally, ViT pretrained with MAE demonstrates superior performance compared to ViT trained with DINOv2.

We hypothesize that, although DINOv2 encoder architecture provides high-level semantic representations, MAE focuses more on fine-grained details, making it more effective for image-goal navigation tasks.

 \begin{table}[h]
\centering
\resizebox{0.5\linewidth}{!}{
\begin{tabular}{l|cc|cc}
\toprule
               & \multicolumn{2}{c|}{Highrise}                                      & \multicolumn{2}{c}{Sanctuary}    \\                    
               & SR         & SPL         & SR         & SPL    \\ \hline
ViT-raw (ZS) &  0.380   &  0.363   &  0.227  &  0.210 \\
ViT-DINOv2 (ZS) &  0.600    &  0.577    &  0.413   & 0.393  \\
ViT-MAE (ZS) &  \textbf{0.693}    &  \textbf{0.64}     &  \textbf{0.540}    &  \textbf{0.503} \\
\hline 
ViT-raw (FT) & 0.860   &  0.803   &   0.613   &  0.567 \\
ViT-DINOv2 (FT) &  0.953   &  0.890  &  0.680  & 0.637 \\
ViT-MAE (FT) &   \textbf{0.967}   & \textbf{0.907}     &   \textbf{0.867}   &  \textbf{0.797} \\
\bottomrule
\end{tabular}
}
\caption{SR and SPL performance on three types of ViT encoders across two environments.  Experiments run for both zero-shot (first two rows) and fine-tuning (last two rows) setting. 
Performances are averaged on three difficulty levels.}
\label{tab:vit}
\end{table}

\paragraph{Effect of Auxiliary Loss.}
We also investigate the effect of each auxiliary loss on the performance of PIG-Nav. The experiments are divided into five different settings. The first only acquires the waypoint information as normal navigation tasks. In the following three settings, we remove the goal, distance and global information separately to show these part effects in auxiliary loss. We show the auxiliary losses are all useful to achieve the success rate of PIG-Nav based on the outcome in Table~\ref{tab:aux}. This experiment also suggests that predicting global decision making information are beneficial to the performance of navigation foundation models. 

 \begin{table}[h]
\centering
\resizebox{0.5\linewidth}{!}{
\begin{tabular}{l|cc|cc}
\toprule
               & \multicolumn{2}{c|}{Highrise}                                      & \multicolumn{2}{c}{Sanctuary}    \\                    
               & SR         & SPL         & SR         & SPL    \\ \hline
Waypoint Only (ZS) &  0.613   &  0.563  &  0.447  & 0.410 \\
No Goal (ZS) &  0.627    &   0.590   &  0.440  & 0.417 \\
No Distance (ZS) &  0.653   & 0.623  &  0.487   & 0.460\\
No Global (ZS) &   0.680  &  0.620 &  0.480   & 0.450\\
All (ZS) & \textbf{0.693}    &  \textbf{0.640}     &  \textbf{0.540}    &  \textbf{0.503} \\ \hline
Waypoint Only (FT) & 0.953    &   0.883 &  0.823  &  0.747 \\
No Goal (FT) &  0.967    &  0.903    &  0.827  & 0.763 \\
No Distance (FT) &  0.960   & 0.900  & 0.840    & 0.763\\
No Global (FT) &  0.953   & 0.897  &  0.840   & 0.750\\
All (FT) &   \textbf{0.967}   & \textbf{0.907}     &   \textbf{0.867}   &  \textbf{0.797} \\
\bottomrule
\end{tabular}
}
\caption{SR and SPL performance with different auxiliary loss designs across two environments. Experiments run for both zero-shot (first two rows) and fine-tuning (last two rows) setting. 
Performances are averaged in three difficulty levels. }
\label{tab:aux}
\end{table}

\section{Conclusions}

In this paper, we introduce PIG-Nav, a novel pretrained image-goal navigation model that advances state-of-the-art visual navigation foundation models from both model design aspect and data aspect. Our early-fusion network, which integrates pretrained Vision Transformer (ViT) encoders, enhances both efficiency and navigation performance. Additionally, we introduce auxiliary loss functions to further optimize pretraining and fine-tuning, improving the model’s adaptability. Furthermore, we demonstrate that newly labeled game video data can serve as a valuable resource for navigation tasks. Our proposed game video labeling pipeline not only enhances dataset quality for PIG-Nav but also holds potential for broader applications, such as Vision-Language-Action (VLA) models for robotic control. 
Through extensive experiments in both simulated game environments and real-world robotic settings, we validate the effectiveness of PIG-Nav. These results highlight its potential to enhance autonomous navigation, paving the way for further advancements in robotic perception, decision-making, and generalizable navigation policies.

An important direction for future research is to assess PIG-Nav’s generalization across more diverse and complex environments, including dynamic and partially observable settings. Additionally, integrating topological mapping could improve the agent’s ability to perform long-horizon navigation and enhance its exploration capabilities in unseen environments. We also hope that the insights presented in this paper may contribute to the design of VLM-based navigation models.

\bibliographystyle{unsrtnat}
\bibliography{references}  

\clearpage
\clearpage

\section{Training Tasks for PIG-Nav}

\paragraph{Notations.} We define the provided navigation trajectory as: 
\begin{equation*}
    \tau=(o_0, o_1, \cdots, o_T; p_0, p_1, \cdots, p_T)
\end{equation*}
where $T$ represents the total number of steps in the trajectory.  The first observation $o_{\text{cur}} = o_0$ corresponds to the current image, while the final observation $o_{\text{goal}} = o_T$ represents the goal image. Each observation $o_k$ is an RGB image that captures the agent’s 90-degree field of view directly in front. Each pose $p_k$ is represented as $(x, y, \psi)$, where $x$ and $y$ denote the agent’s coordinates, and $\psi$ represents the yaw angle, indicating the agent’s orientation.

\paragraph{Waypoint Action Prediction.} 

The goal of waypoint action prediction is to predict the next $N_{\text{waypoint}} = 10$ consecutive steps in the provided navigation trajectory based on the current image observation $o_{\text{cur}}$ and the goal image observation $o_{\text{goal}}$. This chunk-based action prediction approach has been adopted in recent works~\citep{bachmann2024pitfalls, hu2024learning} as an improvement over methods that predict only the next immediate action.

The waypoint action loss is defined as:
\begin{equation*}
    L^{\text{waypoint}}(\tau) = \sum_{k=1}^{N_{\text{waypoint}}}D^{\text{pos\_yaw}}(f_{\theta}^{\text{waypoint}}(o_{cur}, o_{goal})_k, p_{0 \rightarrow k})
\end{equation*}
where the waypoint function $f_{\theta}^{\text{waypoint}}(o_{cur}, o_{goal})_k$ denotes the predicted waypoint action generated by PIG-Nav, and $p_{0 \rightarrow k}$ denotes the pose of $t=k$ relative to the initial pose at $t=0$.  

The following position-yaw $L_2$ metric quantifies the discrepancy between predicted and ground-truth positions and yaw angles, assessing the accuracy of the model’s waypoint predictions.
\begin{align*}
    &D^{\text{pos\_yaw}}(p_1, p_2) = (x_2 - x_1)^2 + (y_2 - y_1)^2 \\
    &\ \ \ + \left(\cos(\psi_2) - \cos(\psi_1)\right)^2 + \left(\sin(\psi_2) - \sin(\psi_1)\right)^2 
\end{align*}

\paragraph{Relative Pose to Goal. }

In the relative pose prediction task, the model estimates the pose of the goal image relative to the observation image. This task enables the model to capture global task information by understanding the spatial relationship between the current observation and the goal image.

The relative position loss function, which measures the discrepancy between the predicted and ground-truth relative positions and orientations, is defined as:
\begin{align*}
    &\ \ \ \ L^{\text{relative}}(o_{cur}, o_{goal};p_0, p_T) \\
    &= D^{\text{pos\_yaw}}(f_{\theta}^{\text{relative}}(o_{cur}, o_{goal}), p_{0\rightarrow T})
\end{align*}
where $f_{\theta}^{\text{relative}}(o_{\text{cur}}, o_{\text{goal}})$ denotes the predicted relative pose, and $p_{0\rightarrow T}$ represents the ground-truth relative pose of the goal image with respect to the initial observation.

\paragraph{Navigation Distance Prediction. }
In the navigation distance prediction task, our model is trained to estimate the total distance required for the agent to navigate from its current state to the goal state. By optimizing this loss, the model learns to infer the connectivity and traversability between different locations in the environment, enabling more efficient path planning.

The navigation distance between the current position and the goal position is defined as:
\begin{equation*}
    \text{nav\_distance} (\tau) = \sum_{k=0}^{T-1}
\left\| 
\begin{pmatrix} 
x_{k+1} - x_k,  y_{k+1} - y_k
\end{pmatrix}
\right\|_2^2
\end{equation*}

The corresponding navigation distance loss function is formulated as:
\begin{equation*}
    L^{\text{nav\_distance}}  (\tau) = (f_{\theta}^{\text{nav\_distance}}(o_{cur}, o_{goal}) -  \text{nav\_distance} (\tau))^2
\end{equation*}

where $f_{\theta}^{\text{nav\_distance}}(o_{cur}, o_{goal})$ represents the model’s predicted navigation distance, and $\text{nav\_distance} (\tau)$ is the ground-truth distance computed from the trajectory.

\paragraph{Global Path Prediction. }
To further enhance the agent’s ability to predict long-range positions, we introduce the global path prediction task. This task requires the model to predict $N_{\text{global}} = 10$ intermediate waypoints, which are equally spaced in time from the current time step to the total trajectory length $T$. These predicted waypoints generated by PIG-Nav are compared to the ground-truth trajectory to compute the loss function.

The corresponding global path prediction loss is defined as:
\begin{equation*}
    L^{\text{global}}(\tau) = \sum_{k=1}^{N_{\text{global}}}D^{\text{pos\_yaw}}(f_{\theta}^{\text{global}}(o_{cur}, o_{goal})_k, p_{0 \rightarrow\lfloor\frac{k \times T}{N_{\text{global}}}\rfloor})
\end{equation*}

where $f_{\theta}^{\text{global}}(o_{\text{cur}}, o_{\text{goal}})_k$ represents the predicted waypoint at step $k$, and $p_{0 \rightarrow \lfloor\frac{k \times T}{N_{\text{global}}}\rfloor}$ denotes the corresponding ground-truth waypoint in the trajectory.

\section{Pretraining Dataset}
\begin{figure*}[htb]
\centerline{\includegraphics[scale=0.45]{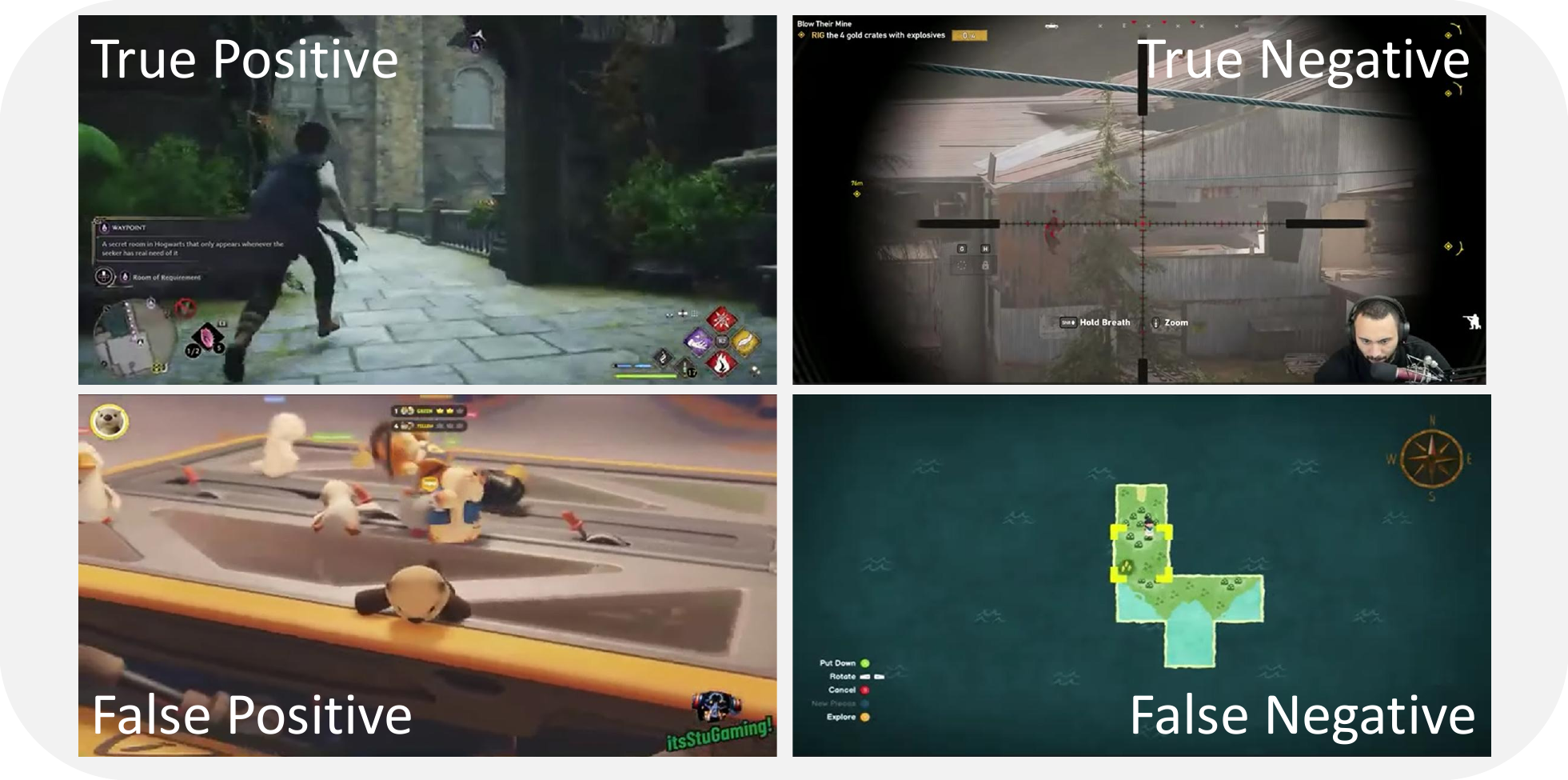}}
\caption{Examples of Qwen2 two steps filtering.}
\label{fig:10}
\end{figure*}
\label{method}

\begin{figure*}[htb]
\centerline{\includegraphics[scale=0.5]{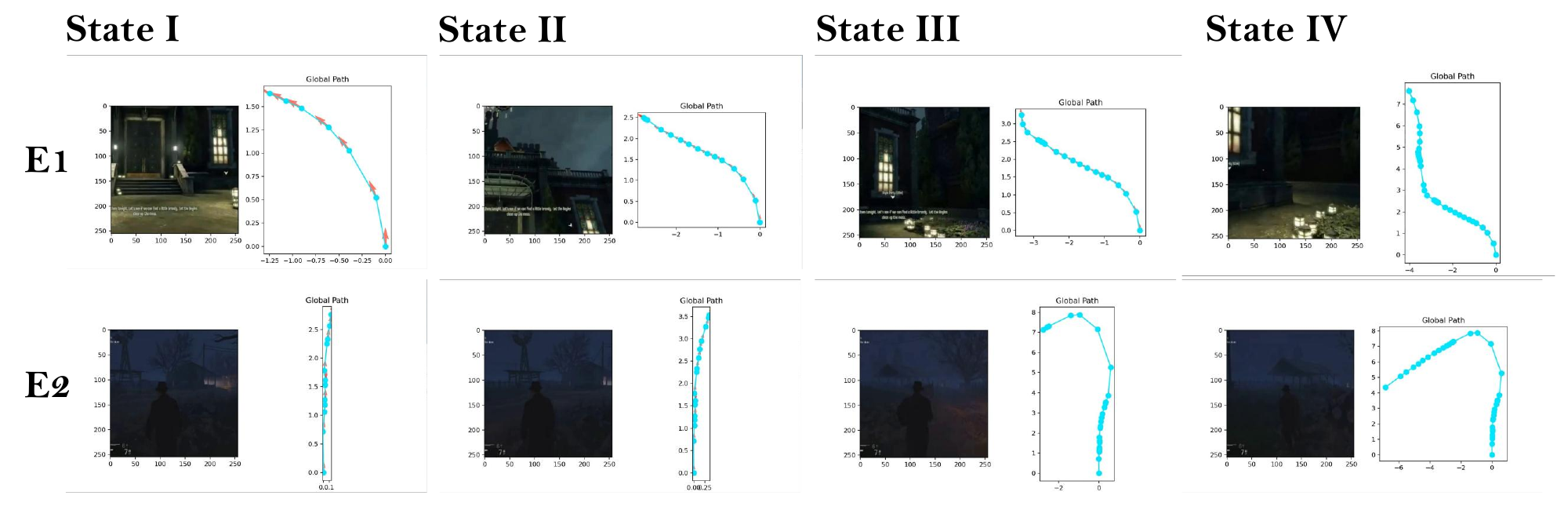}}
\caption{Visualization of IDM labeled data. Four stages from Stage I to Stage II for two environment E1 and E2 are listed in sequence.}
\label{fig:11}
\end{figure*}
\label{method}

\subsection{Open Public Navigation Datasets}
While we incorporate publicly available datasets, certain data sources remain unavailable compared to those used in \citep{shah2023gnm, shah2023vint, sridhar2024nomad}. However, despite being pretrained on a smaller dataset, our model outperforms existing approaches, demonstrating its effectiveness and robustness.

For existing datasets, two key aspects require careful consideration. First, data cleaning: we identified noise in some datasets, such as abnormally large position changes between consecutive time steps. To ensure data quality and facilitate stable training, we filtered out data points where position changes exceeded five times the average displacement. Second, data normalization: since datasets vary in scale and properties, transforming them into a unified space is crucial for effective training. We experimented with three normalization techniques and selected the one that yielded the best performance. 

\subsection{Labeling Game Video Data}

\paragraph{Filtering.}
We show the final classification outcome of our two steps filtering by Qwen2~\citep{wang2024qwen2} VLM in Figure~\ref{fig:10}. In details, Figure~\ref{fig:10} show the four types of Qwen2 model classification outputs, including True Positive, False Negative, True Negative and False Negative separately.

Additionally, in the first step filtering, we focus on our navigation tasks and use the prompt \textit{"If this video can be used in the construction of navigation task dataset, then return True, else return False."} For the second step filtering, we test total 12 prompts in Qwen2 with our labeling 500 positive samples, which are the video segments with True label by Qwen2 in classification. The prompt \textit{"If this video does not contain the interaction with game menu and this video measures the movement, then return True, else return False."} performs best in our criterion which   combines 80\% precision and 20\% recall as the final selection metric.

\paragraph{IDM Architecture.}
 For data labeling, We utilize the ViT model architecture as PIG-Nav to encode the change of two consecutive frames and construct an attention head to decode the navigation action to vector representations $[\Delta x, \Delta y, \cos \Delta\psi, \sin\Delta\psi]$.
 We show the labeled output of our IDM in Figure~\ref{fig:11}.

 We also propose to leverage pre-trained latent action representation encoders based on our final raw video segments firstly to enhance the model abilities that it measures the diverse videos. Then we fine-tune this latent action representation encoders with a 3 layers MLP as its Inverse Dynamics Model to build the relationship between these latent action representations and the low level action representations. We highlight that the pre-trained latent action encoder can capture the similarity among similar action motion in video segments and use the high dimension latent action vector to represent it. We find this structure needs a precise IDM model design to translate the latent action, representing as the high dimensional vector, to the low level action as the navigation action vector.

\section{Rollout Examples in Game Environment}

\begin{figure*}[htb]
\centerline{\includegraphics[scale=0.4]{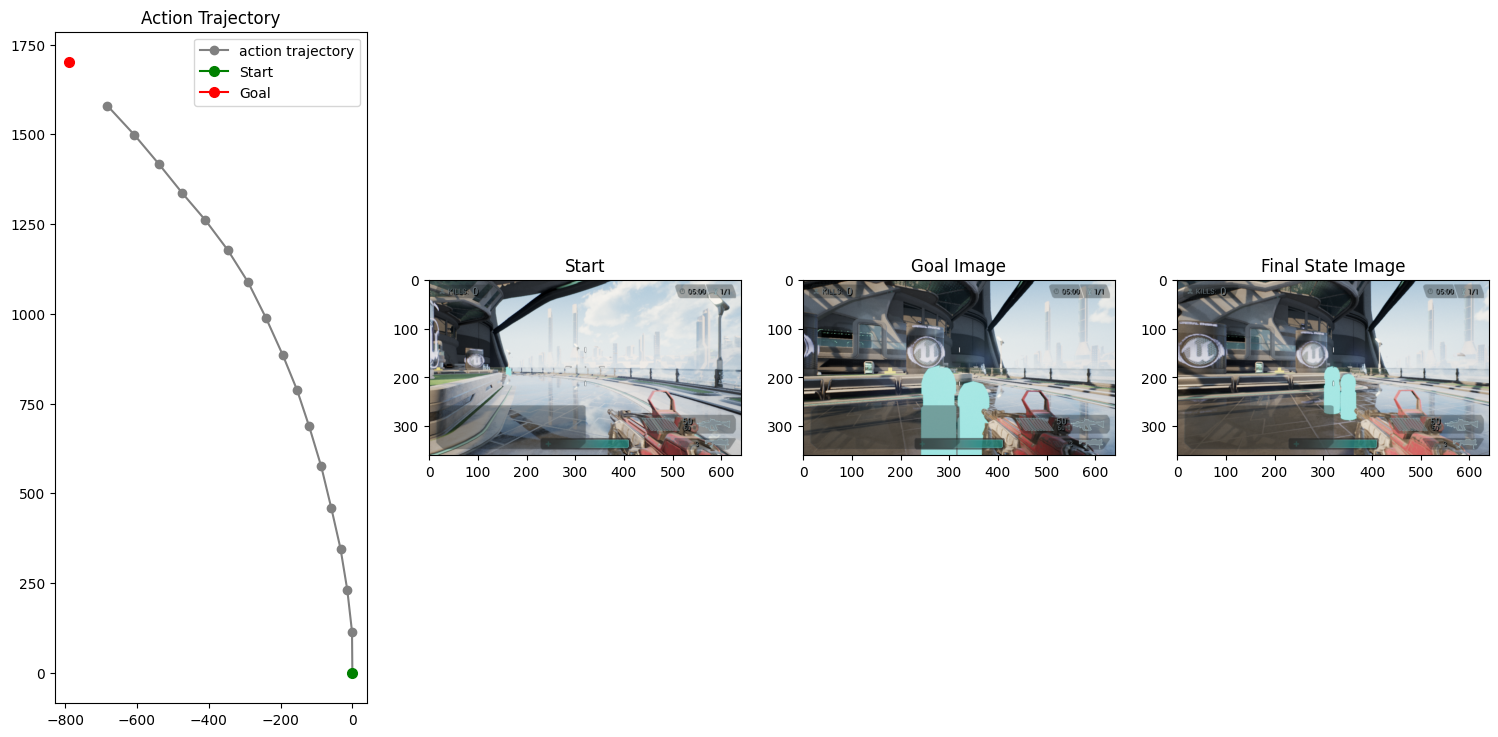}}
\caption{Rollout example in Highrise environment.}
\label{fig:casestudy-highrise}
\end{figure*}

\begin{figure*}[htb]
\centerline{\includegraphics[scale=0.4]{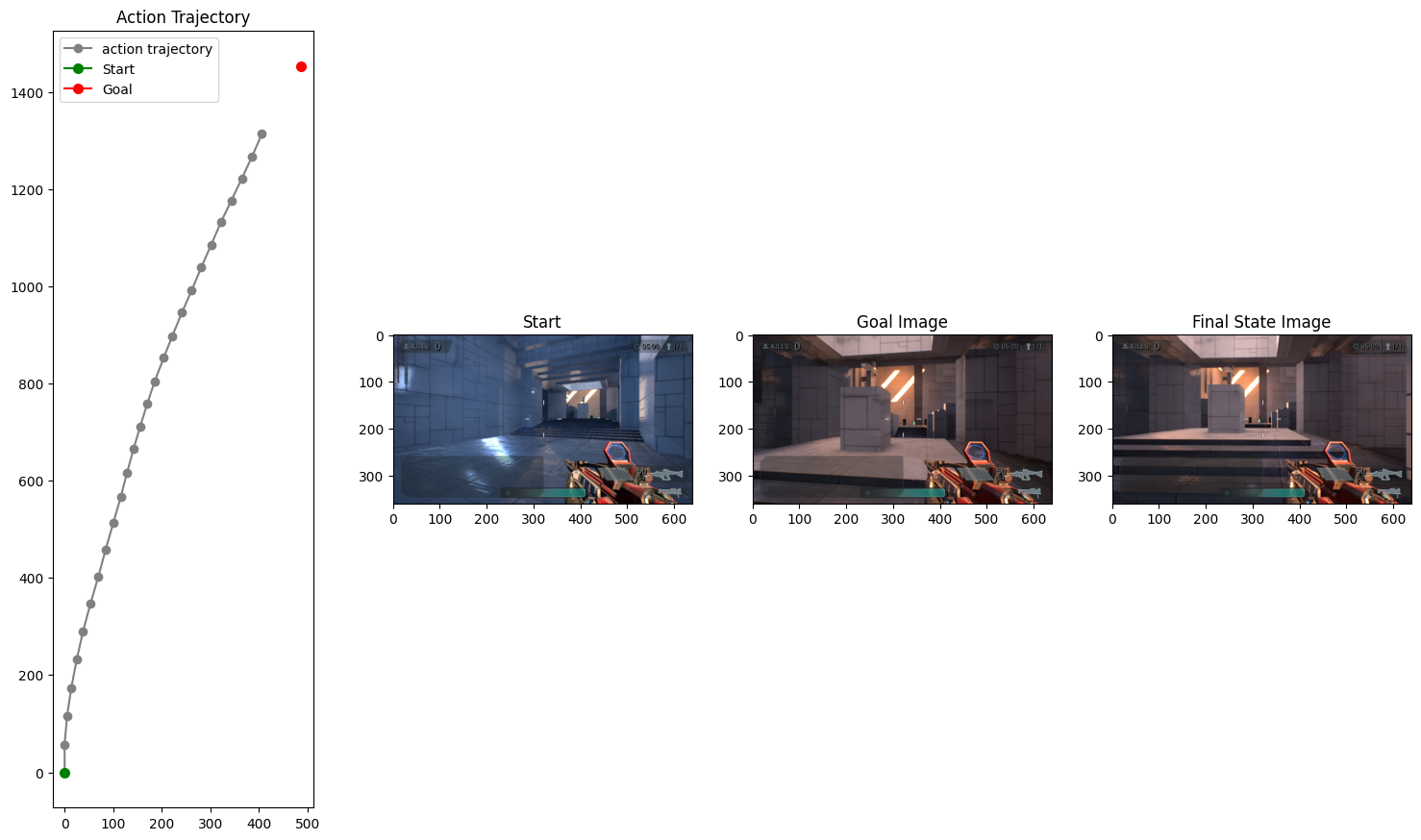}}
\caption{Rollout example in Sanctuary environment.}
\label{fig:casestudy-sanctuary}
\end{figure*}

In Figure~\ref{fig:casestudy-highrise} and \ref{fig:casestudy-sanctuary}, we show rollout examples in two large-scale game envionment: Highrise and Sanctuary, where the start and goal position is sampled from the validation set. The observation at the start point, goal image provided by the PIG-Nav, final observation at the end of the execution, and the 2D trajectory of PIG-Nav are shown in the figure. PIG-Nav successfully navigate to the target position with only current and goal image as inputs. 

\section{Ablation Study in Details}
 We explain our ablation studies in details for three difficulty settings in Table~\ref{tab:appendix_ablation_earlyfuse}, \ref{tab:appendix_ablation_vit} and \ref{tab:appendix_ablation_aux}.

 Table~\ref{tab:appendix_ablation_earlyfuse} measures the effectiveness of our proposed early-fusion structure, which performs better than non-fuse in all difficulty settings for both Highrise and Sanctuary environments.

 \begin{table}[h]
\centering
\resizebox{1.0\linewidth}{!}{
\begin{tabular}{l|cccccc|cccccc}
\toprule
               & \multicolumn{6}{c|}{Highrise}                                      & \multicolumn{6}{c}{Sanctuary}    \\                                 
               & \multicolumn{2}{c}{Easy} & \multicolumn{2}{c}{Medium} & \multicolumn{2}{c|}{Hard} & \multicolumn{2}{c}{Easy} & \multicolumn{2}{c}{Medium} & \multicolumn{2}{c}{Hard} \\
               & SR         & SPL         & SR         & SPL         & SR         & SPL        & SR         & SPL         & SR         & SPL         & SR         & SPL   \\ \hline
Non-Fuse (ZS) &     0.74     &     0.69    &  0.48          &       0.45     &      0.24      &     0.19       &   0.56         &    0.44         &       0.20     &      0.15       &      0.08       & 0.06        \\
Early Fuse (ZS) &     \textbf{0.90}        &     \textbf{0.86}        &  \textbf{0.72}          &    \textbf{0.68}         &          \textbf{0.46}  &     \textbf{0.38}       &   \textbf{0.84}         &      \textbf{0.80}       &      \textbf{0.48}      &      \textbf{0.45}      &   \textbf{0.30} & \textbf{0.26}                  \\ \hline
Non-Fuse (FT) &     \textbf{1.00}       &      0.96       &   0.94         &    0.88         &     0.84       &     0.74       &   \textbf{1.00}    &  0.94   &      \textbf{0.84}       &     \textbf{0.75}       &     0.72        & 0.62                    \\ 
Early Fuse (FT) &     \textbf{1.00}        &      \textbf{ 0.97}      &  \textbf{1.00}          &      \textbf{0.94}       &        \textbf{0.90}    &     \textbf{0.81}       &   \textbf{1.00}        &   \textbf{0.96}          &    \textbf{0.84}        & \textbf{0.75}            &    \textbf{0.76} & \textbf{0.68 }                \\ 
\bottomrule
\end{tabular}
}
\caption{SR and SPL performance across two environments and three difficulty levels with different model structure. Experiments run for zero-shot (first two rows) and fine-tuning (last two rows) setting. Two types of Model structure include non-early fuse and early fuse. All values are rounded to the nearest hundredth.}
\label{tab:appendix_ablation_earlyfuse}
\end{table}

In Table~\ref{tab:appendix_ablation_vit}, we test different ViT encoder structures and find the MAE structure performs best in all three difficulty settings for both Highrise and Sanctuary environments.

\begin{table}[h]
\centering
\resizebox{1.0\linewidth}{!}{
\begin{tabular}{l|cccccc|cccccc}
\toprule
               & \multicolumn{6}{c|}{Highrise}                                      & \multicolumn{6}{c}{Sanctuary}    \\                                 
               & \multicolumn{2}{c}{Easy} & \multicolumn{2}{c}{Medium} & \multicolumn{2}{c|}{Hard} & \multicolumn{2}{c}{Easy} & \multicolumn{2}{c}{Medium} & \multicolumn{2}{c}{Hard} \\
               & SR         & SPL         & SR         & SPL         & SR         & SPL        & SR         & SPL         & SR         & SPL         & SR         & SPL   \\ \hline
ViT-raw (ZS)  &    0.76        &   0.73          &   0.26         &      0.25       &        0.12    &       0.11    &   0.46         &   0.43   &   0.14    &    0.13        &      0.08       &        0.07             \\
ViT-DINOv2 (ZS) &   0.82         &  0.81           &        0.70    &     0.67        &    0.28        &     0.25       &         0.70   &       0.67      &   0.32         &         0.30    &     0.22     &     0.21      \\
ViT-MAE (ZS) &        \textbf{0.90}        &     \textbf{0.86}        &  \textbf{0.72}          &    \textbf{0.68}         &          \textbf{0.46}  &     \textbf{0.38}       &   \textbf{0.84}         &      \textbf{0.80}       &      \textbf{0.48}      &      \textbf{0.45}      &   \textbf{0.30} & \textbf{0.26}              \\ \hline
ViT-raw (FT)  &   \textbf{1.00}         &   0.96          &     0.86       &         0.80    &   0.72         &     0.65       &   0.88         &   0.83          &     0.58       &     0.53        &  0.38 & 0.34                   \\
ViT-DINOv2 (FT) &   \textbf{1.00}         &   0.97          &       \textbf{1.00}     &       \textbf{0.94}      &       0.86     &    0.76        &    0.96        &   0.92          &   0.62         & 0.58            &   0.46 &0.41\\ 
ViT-MAE (FT) & \textbf{1.00}        &       \textbf{0.97}      &  \textbf{1.00}          &      \textbf{0.94}       &        \textbf{0.90}    &     \textbf{0.81}       &   \textbf{1.00}         &   \textbf{0.96}          &    \textbf{0.84}        & \textbf{0.75}            &    \textbf{0.76} & \textbf{0.68}                 \\ 
\bottomrule

\end{tabular}
}
\caption{SR and SPL performance across two environments and three difficulty levels with different pretrained model encoder. Experiments run for both zero-shot (first two rows) and fine-tuning (last two rows) setting. Three model encoders including ViT-raw, ViT-DINOv2 and ViT-MAE apply in experiments. All values are rounded to the nearest hundredth.}
\label{tab:appendix_ablation_vit}
\end{table}

 We show the effectiveness of our auxiliary loss design in Table~\ref{tab:appendix_ablation_aux}. Five settings including waypoint only, no goal, no distance, no global and all auxiliary supports are tested for both Highrise and Sanctuary environments.

\begin{table}[h]
\centering
\resizebox{1.0\linewidth}{!}{
\begin{tabular}{l|cccccc|cccccc}
\toprule
               & \multicolumn{6}{c|}{Highrise}                                      & \multicolumn{6}{c}{Sanctuary}    \\                                 
               & \multicolumn{2}{c}{Easy} & \multicolumn{2}{c}{Medium} & \multicolumn{2}{c|}{Hard} & \multicolumn{2}{c}{Easy} & \multicolumn{2}{c}{Medium} & \multicolumn{2}{c}{Hard} \\
               & SR         & SPL         & SR         & SPL         & SR         & SPL        & SR         & SPL         & SR         & SPL         & SR         & SPL   \\ \hline
Waypoint Only (ZS) &     0.82     &  0.77         &       0.70     &    0.64        &    0.32        &       0.28     & 0.78           &     0.72        &    0.42        &     0.38        &     0.14 & 0.13                \\
No Goal (ZS) &    0.82        &       0.79      &     0.66      &     0.63        &         0.40   &  0.35          &    0.78        &    0.76         &        0.38    &     0.35        &   0.16 & 0.14                  \\
No Distance (ZS) &    0.86        &     0.84        &  0.72          &    \textbf{0.69}         &      0.38      &   0.34         &   0.82         &   0.79          &     0.46       &      0.43       &   0.18 & 0.16                  \\
No Global (ZS) &   0.88         &       0.84      &    \textbf{0.74}        &       0.68      &        0.42    &       0.34     &   0.80        &  0.76           &      0.46      &     0.42        &   0.18 & 0.17                  \\ 
All (ZS)    &     \textbf{0.90}        &      \textbf{0.86}        &   0.72          &     0.68         &           \textbf{0.46}  &      \textbf{0.38}       &    \textbf{0.84}         &       \textbf{0.80}       &       \textbf{0.48}      &       \textbf{0.45}      &    \textbf{0.30} &  \textbf{0.26}                   \\ \hline
Waypoint Only (FT) &    \textbf{1.00}        &  0.96           &     \textbf{1.00}       &       0.93      &         0.86   &     0.76       &     \textbf{1.00}       &     0.95        &    \textbf{0.84}        &  0.73           &    0.66 & 0.56                 \\ 
No Goal (FT) &   \textbf{1.00}         &      \textbf{0.97}       &     \textbf{ 1.00}      &     0.93        &   \textbf{0.90}         &    \textbf{0.81}        &      \textbf{1.00}      &         \textbf{0.96}    &      0.82      & 0.74            &   0.66    &0.59              \\ 
No Distance (FT)  &    \textbf{1.00}        &   \textbf{0.97}          &  \textbf{1.00}         &     \textbf{0.94}        &        0.88    &     0.79       &    \textbf{1.00}        &      \textbf{0.96}       &    \textbf{0.84}        & 0.74            &      0.68 & 0.59               \\ 
No Global (FT) &     \textbf{1.00}       &    \textbf{0.97}         &       \textbf{1.00}     &    \textbf{0.94}         &  0.86          &  0.78          &   \textbf{1.00}         &        0.95     &   \textbf{0.84}         &   0.72          &  0.68                &0.58   \\ 
All (FT)  & \textbf{1.00}        &       \textbf{0.97}      &  \textbf{1.00}          &      \textbf{0.94}       &        \textbf{0.90}    &     \textbf{0.81}       &   \textbf{1.00}         &   \textbf{0.96}          &    \textbf{0.84}        & \textbf{0.75}            &    \textbf{0.76} & \textbf{0.68}                 \\ 
\bottomrule
\end{tabular}
}
\caption{SR and SPL performance across two environments and three difficulty levels with different auxiliary loss design. Experiments run for both zero-shot (first two rows) and fine-tuning (last two rows) setting. Five types of design includes waypoint only, without goal, without distance, without global and all. All values are rounded to the nearest hundredth.}
\label{tab:appendix_ablation_aux}
\end{table}






\end{document}